# Identifying TBI Physiological States by Clustering Multivariate Clinical Time-Series Data

Hamid Ghaderi[1], Brandon Foreman, MD[2], Amin Nayebi[1], Sindhu Tipirneni[3], Chandan K. Reddy, Ph.D.[3], Vignesh Subbian, Ph.D.[1]

[1] College of Engineering, University of Arizona, Tucson, AZ, USA; [2] College of Medicine, University of Cincinnati, Cincinnati, OH, USA; [3] Department of Computer Science, Virginia Tech, Arlington, VA, USA

**Abstract**

*Determining clinically relevant physiological states from multivariate time-series data with missing values is essential for providing appropriate treatment for acute conditions such as Traumatic Brain Injury (TBI), respiratory failure, and heart failure. Utilizing non-temporal clustering or data imputation and aggregation techniques may lead to loss of valuable information and biased analyses. In our study, we apply the SLAC-Time algorithm, an innovative self-supervision-based approach that maintains data integrity by avoiding imputation or aggregation, offering a more useful representation of acute patient states. By using SLAC-Time to cluster data in a large research dataset, we identified three distinct TBI physiological states and their specific feature profiles. We employed various clustering evaluation metrics and incorporated input from a clinical domain expert to validate and interpret the identified physiological states. Further, we discovered how specific clinical events and interventions can influence patient states and state transitions.*

**Introduction**

Traumatic Brain Injury (TBI) is a major contributor to death and disability, especially among young adults [1]. Globally, an alarming estimate of 69 million individuals suffer from TBI annually [2], rendering it a pertinent public health issue. The negative consequences of TBI can be severe, resulting in physical, cognitive, and psychological impairments that can last for a long time and affect an individual's quality of life. As a result, it is important to understand the underlying causes of TBI and develop effective treatments. To successfully manage severe TBI, it is crucial to quickly detect signs of an energy crisis and provide appropriate physiological support to prevent further damage to the brain [3]. Conventionally, sporadic serial examinations of the neurological state and imaging are utilized to detect secondary brain injury. However, such assessments provide only a limited glimpse of the injury, and they primarily reflect the damage that has already taken place. On the other hand, real-time data acquired from bedside monitors has the potential to facilitate prompt and preemptive identification of secondary brain injury [4,5]. However, the multitude of data supplied by multimodal physiological monitoring provides an overwhelming amount of data for clinicians, making it difficult to comprehend and integrate meaningful information in a uniform and efficient manner [6]. Consequently, clinicians are inevitably compelled to select certain variables to monitor based on their experience. Moreover, the current guidelines for treatment based on monitoring mainly concentrate on individual parameters, such as blood pressure or intracranial pressure (ICP), which leads to simplified treatment strategies for conditions with complex, varied, and interdependent pathophysiological processes that change dynamically over time after an injury [7–9]. For instance, intracranial hypertension can arise due to various mechanisms, each requiring distinct therapies and pose different risks for brain damage. However, ICP management protocols view elevated ICP as a single problem and suggest identical therapies without considering the specific underlying cause [10].

*Significance of Identifying Physiological States:* TBI is a prime example of an acute, heterogenous condition that results in various patterns of physiological states among patients. Identifying the physiological states of TBI patients is crucial to comprehending their present health status and administering appropriate medical intervention [11]. Clinicians must monitor patients' physiological states over time to accurately assess their condition and identify optimal treatment strategies. Monitoring physiological states over time enables us to recognize patterns and variations from the anticipated path of recovery and prioritize patients who may display major physiological irregularities, ensuring prompt intervention. Treatment options for TBI carry differing risk-benefit profiles, necessitating careful consideration of individual patient characteristics when selecting the best course of action [11,12]. Therefore, adopting a personalized approach to TBI management, founded on precise and timely evaluations of physiological states, is crucial to achieving optimal outcomes.

*Clustering methods and time-series physiologic data:* Physiological data are often represented as time-series variables that are measured at irregular time intervals and may contain missing values. Clustering techniques are generally promising for subgrouping samples within heterogeneous clinical datasets related to a wide range of diseases and injuries, offering valuable insights for research and treatment. However, there has been a lack of emphasis on utilizing clustering methods specifically designed for time-series data. Instead, studies often rely on clustering methods suitable for less complex data types such as aggregated measurements or fixed time point data [13]. These methods may not fully capture the temporal variability inherent in physiological processes, potentially overlooking important information. Thus, there is a need to apply clustering approaches that can effectively handle the complexity and dynamic nature of multivariate time-series physiological data.

*Integration of self-supervision and clustering:* Clustering methods for multivariate time-series data fall under two primary categories: the methods based on raw data and those utilizing representations [14]. The former can be negatively impacted by noise and outlier data, whereas the latter employs the representations of the input data, mitigating such issues [14]. Deep learning models can enhance the effectiveness of representation-based clustering. However, these models require large, labeled datasets that might not be easily obtainable in most medical settings [15]. There is an emerging trend toward the use of self-supervised learning for the clustering of multivariate time-series data. This process entails training a deep learning model on an unlabeled time-series dataset to learn the representations. These representations are then subsequently employed for various clustering tasks [16]. However, the current self-supervised learning-based methods are only effective when there are no missing values, which is rarely the case. There are several commonly employed techniques to address the issue of missing values in data, including the omission of incomplete samples, the aggregation of data into discrete time periods, the imputation of missing values, and data interpolation [16]. However, each of these methods has inherent limitations that may result in biased analyses, invalid conclusions, or a loss of important, fine-grained information. The Self-supervised Learning-based Approach to Clustering multivariate Time-series data with missing values (SLAC-Time) is a novel clustering approach based on Transformers that treats each multivariate time-series data as a set of observation triplets, which eliminates the need to handle missing values [16]. This method leverages unlabeled data by using time-series forecasting as a proxy task to learn robust time-series representations. The approach simultaneously learns the neural network parameters and the cluster assignments of the representations it has learned. To enhance the model's performance, it repeatedly clusters the learned representations using the K-means method and employs the resulting cluster assignments from each iteration as pseudo-labels to update the parameters of the model (Figure 1).

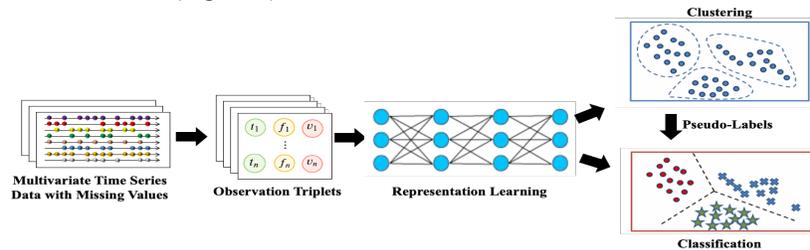

**Figure 1.** Architecture of SLAC-Time

In this study, we aim to fill the gaps in the current literature by applying the SLAC-Time algorithm to cluster multivariate time-series physiological data with missing values. Unlike previous studies, we do not rely on data imputation or aggregation methods, allowing us to preserve the integrity of the original data. By leveraging input from clinical domain experts (e.g., trauma intensivists) and using a variety of cluster validation techniques, we seek to identify distinct TBI physiological states and their associated feature profiles. Additionally, we investigate how physiological states change in response to specific clinical events among severe TBI patients.

**Related Work**
This literature review examines the application of clustering techniques to physiological data, based on analysis of six relevant studies. Each of the studies described below focuses on the potential benefits of clustering analysis in identifying physiological states that can inform clinical decision-making. Clustering methods in these studies include hierarchical clustering, k-means clustering, subspace clustering, and hidden Markov model.

Schrumpf et al. [17] proposed an iterative hierarchical clustering technique to group physiological data into health states based on their similarity. The authors tested their approach on electrocardiogram (ECG) data from individuals experiencing different levels of physical strain and discovered a strong association between the identified health states and the experimental phases of the exercise. Bidargaddi et al. [18] investigated the ability to differentiate between normal and abnormal physiological states in cardiac patients using k-means clustering on heart rate (HR), HR variability (HRV), and movement activity data. They found that two clusters in the inactive state with high HR and low HRV or low HRV and low HR were associated with pathological states, whereas activity information was significant in distinguishing between normal and abnormal clusters found in the inactive and active states with low HRV. Rajagopalan et al. [19] used hierarchical cluster analysis to group hourly values of physiological data, including HR, mean arterial pressure, ICP, brain tissue oxygen, and cerebral microdialysis, obtained from patients with severe TBI or high-grade subarachnoid hemorrhage. They identified four clusters with distinct physiological profiles that corresponded to known physiological states observed following acute brain injury. The distribution of these clusters was then compared between patients who had favorable and unfavorable outcomes. The study found that patients with favorable outcomes had a higher proportion of normal physiological states, while patients with unfavorable outcomes had a higher proportion of ischemia and hyperglycolysis states.

Ehsani et al. [20] conducted a study using subspace clustering methods to cluster time-series data collected from TBI patients. They focused on density-based, cell-based, and clustering-oriented methods and identified three clusters of laboratory physiological data that were associated with mortality status. They found that the density-based algorithm had the highest F1 score and coverage in grouping patients based on mortality status. Cohen et al. [21] proposed a hierarchical agglomerative clustering algorithm for identifying subgroups of patients with acute respiratory distress syndrome (ARDS) based on non-temporal physiological and demographic data. They found that their clustering approach could identify groups of ARDS patients with distinct physiological profiles and survival rates. Asgari et al. [13] developed a method to identify physiological states in TBI patients by applying a hidden Markov model to hourly physiological data. They found that clustering using a hidden Markov model can simplify a complex set of physiological variables into a clinically plausible sequence of time-sensitive physiological states while retaining prognostic information. They discovered three physiological states based on various measurements such as ICP, cerebral perfusion pressure (CPP), compensatory reserve index, and autoregulation status. These states were then categorized as "good," "intermediate," or "poor" using a three-part state variable.

While these studies demonstrate the potential utility of clustering techniques for physiological data, their focus is on clustering non-temporal physiological data obtained from multivariate time-series datasets. As a result, they may not fully capture the temporal variability inherent in physiological processes, potentially overlooking important information. Therefore, there is a need to apply clustering approaches that can effectively handle the complexity and dynamic nature of multivariate time-series physiological data.

**Materials and Methods**
*Data Source*
Our study utilized data from the Transforming Research and Clinical Knowledge in Traumatic Brain Injury (TRACK-TBI) dataset, comprising of detailed clinical information on 2996 TBI patients with various severity levels, collected from 18 academic Level I trauma hospitals across the United States [22]. Specifically, our investigation focused on 16 TBI patients for whom high-resolution recordings of physiological data were available. Each TBI patient record was represented as a multivariate time-series dataset, consisting of eight variables: diastolic blood pressure (DBP), systolic blood pressure (SBP), CPP, HR, ICP, pressure reactivity index (PRx), respiratory rate (RR), and oxygen saturation ($SpO_2$). The length of the time-series variables varied across patients, ranging from two to 179 hours. We divided each multivariate time-series data into multiple non-overlapping samples of equal length that are clinically relevant. We chose a length of 30 minutes for each sample, as this is the typical interval for interventions in ICUs. This process yielded a total of 2919 samples for analysis. Additionally, within each sample, we calculated the average value of each variable at 10-second intervals, resulting in 180 time steps within each 30-minute sample. Notably, despite these computations, the samples include missing values. It is important to note that the identifiers used for the TBI patients in this study are randomly generated and do not correspond to actual patient identifiers, thus maintaining patient confidentiality and abiding by data privacy regulations. All patients were also enrolled in the TRACK-TBI study under informed consent from the patient's legal authorized representative and approval of local institutional review boards at each site.

*Clustering Approach*

In this study, we use the SLAC-Time approach proposed by Ghaderi et al. [16] for clustering multivariate time-series physiological data. SLAC-Time uses a self-supervised Transformer model for time series (STraTS) to learn representations of multivariate time-series data [23]. To deal with missing data and irregularities in multivariate time-series data, unlike conventional techniques that consider each multivariate time-series data as a matrix with defined dimensions, SLAC-Time regards each multivariate time-series data as a collection of observation triplets. These triplets consist of three components: time of measurement ($t \in \mathbb{R}_{\geq 0}$), variable name ($f$), and variable value ($v \in \mathbb{R}$). Therefore, this technique helps us circumvent the requirement for data aggregation and imputation methods. The Transformer-based architecture of STraTS enables parallel processing of observation triplets. Moreover, the observation triplets are embedded using the Continuous Value Embedding method, which eliminates the requirement for categorizing continuous values before embedding. We use $f_\theta$ with $\theta$ representing the model's parameters to denote the STraTS mapping. To determine the optimal parameter $\theta^*$ that results in general-purpose representations, we consider a training set of N unlabeled samples denoted by $\mathcal{D} = \{T^k\}_{k=1}^{N}$, where the $k^{th}$ sample is a multivariate time-series data $T^k$.

SLAC-Time is a clustering approach which uses self-supervised learning to extract pseudo-labels from the learned representations of the input data. The algorithm is composed of three modules: self-supervision, pseudo-label extraction, and classification.

*Self-supervision module*: In this module, the algorithm uses a larger dataset $N' \geq N$ for time-series forecasting as a self-supervision task. We address missing values in the forecast outputs by implementing a masked Mean Squared Error (MSE) loss during the training of the forecast model. The self-supervision loss is given by Equation 1, which includes the forecast mask ($m^k \in \mathbb{R}$) that serves as an indicator of whether a variable has been recorded in the forecast window, the related values of the variable ($z^k \in \mathbb{R}$), and the forecast output ($\tilde{z}^k$). The ground truth forecast ($z_j^k$) for the $j^{th}$ variable in the $k^{th}$ sample is considered unavailable if the forecast mask is 0, while it is considered available if the forecast mask is 1.

$$\mathcal{L}_{ss} = \frac{1}{|N'|} \sum_{k=1}^{N'} \sum_{j=1}^{|\mathcal{F}|} m_j^k \left( \tilde{z}_j^k - z_j^k \right)^2 \qquad (1)$$

*Pseudo-label extraction module*: Once the STraTS model is pretrained via executing the forecast task, the final layer specifically associated with the forecast task is detached. Subsequently, this adjusted model is employed to compute the non-temporal and time-series embeddings of the samples. Then, K-means clustering is applied on the learned representations. The input for K-means is the learned representations, and the output is k subgroups, learned through its geometric criterion. K-means learns a centroid matrix C of size $d \times k$ and the cluster assignment $y_n$ of each sample $n$, following the equation:

$$\min_{C \in \mathbb{R}^{d \times k}} \frac{1}{N} \sum_{n=1}^{N} \min_{y_n \in \{0,1\}^k} \| f_\theta(T^n) - C y_n \|_2^2 \text{ such that } y_n^\top 1_k = 1 \qquad (2)$$

The optimal cluster assignments $(y_n^*)_{n \leq N}$ are obtained by optimizing this problem. These optimal assignments are considered as pseudo-labels of the samples. Hence, each sample ($T^n$) has a pseudo-label $y_n$ in $\{0,1\}^k$, indicating the sample's membership in one of the $k$ possible predefined classes.

*Classification module*: The acquired pseudo-labels are utilized to supervise the training of a classifier $g_W$ which predicts accurate label based on $f_\theta(T^n)$ representations from the STraTS model. Subsequently, the parameters W of the classifier and the parameters θ of the STraTS model are learned concurrently through the optimization problem below:

$$\min_{\theta, W} \frac{1}{N} \sum_{n=1}^{N} \ell\left(g_W(f_\theta(T^n)), y_n\right) \qquad (3)$$

where $\ell$ denotes a cross-entropy loss function. SLAC-Time is an iterative process consisting of two alternating modules: (1) clustering the representations using the K-means algorithm and subsequently generating pseudo-labels based on cluster assignments, and (2) updating the parameters of both the classifier and the STraTS model to predict accurate labels for each sample by minimizing the loss function.

*Implementation Details*

To pre-train our model and acquire a preliminary representation of multivariate time-series data, we use time-series forecasting as a proxy task. To accomplish this, we select a set of observation windows, including 20, 40, 80, 160, and 175 time steps, followed by a prediction window spanning 5 time steps immediately succeeding the respective observation window. We split our data into training and validation datasets, with a ratio of 80:20. In this study, the target task of SLAC-Time is to subgroup 30-minute samples based on eight time-series features. For this task, we split these samples into training and validation datasets, with the same 80:20 ratio as our proxy task.

We developed SLAC-Time using Keras and TensorFlow backend with the following hyperparameters: 100 output neurons in feed-forward networks, 2 blocks in the Transformer, 4 attention heads in multi-head attention layers, a dropout rate of 0.2, and a learning rate of 5e-4. We trained both proxy and target task models using a batch size of 8 and the Adam optimizer. For the proxy task, we terminated training if there was no reduction in validation loss for ten consecutive epochs. Similarly, we trained the target task for 500 iterations, each of which had 200 epochs, and stopped the training process for each iteration if there was no reduction in validation loss for ten consecutive epochs. Our experiments were conducted on an NVIDIA Tesla P100 GPU, which took nearly three days to finish. The implementation code is publicly available in our GitHub repository at https://github.com/vsubbian/SLAC-Time.

*Data Analysis*

We utilized various intrinsic measures for clustering evaluation to assess the excellence of the clusters across different cluster numbers. These measures include the Silhouette coefficient, which measures the similarity of objects within a cluster compared to objects in neighboring clusters with values ranging from -1 to 1, where a higher value indicates better clustering. The Calinski Harabasz index is also used, calculating the ratio of between-cluster variance to within-cluster variance, where higher values indicate better cluster separation. Additionally, the Dunn index computes the ratio of the smallest inter-cluster distance to the largest intra-cluster distance, with higher values suggesting better clustering. Finally, the Davies Bouldin index estimates the average similarity between each cluster and its most similar cluster, with lower values signifying better cluster separation. We conducted univariate analyses of the mean of time-series variables of physiological states and utilized the Kruskal-Wallis test to determine whether significant differences exist between the physiological states. We consider differences to be significant when the corresponding p-values are less than 0.05. In addition, a TBI clinician evaluated the validity of physiological states and provided interpretations accordingly.

**Results**

We evaluated cluster counts of 3, 4, and 5 and found that using a value of k=3 yields the best clustering performance for all four clustering evaluation metrics (Table 1), indicating that there may be three distinct physiological states. Using k=3 as the number of physiological states, we performed SLAC-Time for clustering 2919 samples, which resulted in three physiological states A, B, and C that include 987, 910, and 1022 samples, respectively. Figure 2 illustrates the physiological states of 16 TBI patients over time. Except for Patient 13, each patient demonstrates a predominant physiological state despite occasional fluctuations caused by clinical events or other medical factors.

Table 1. SLAC-Time performance for different numbers of clusters

| # clusters | Silhouette coefficient | Dunn index | Davies Bouldin index | Calinski Harabasz index |
|---|---|---|---|---|
| 3 | 0.1 | 0.1 | 2.5 | 74.2 |
| 4 | 0.08 | 0.09 | 4.2 | 62.5 |
| 5 | 0.07 | 0.07 | 6.7 | 54.7 |

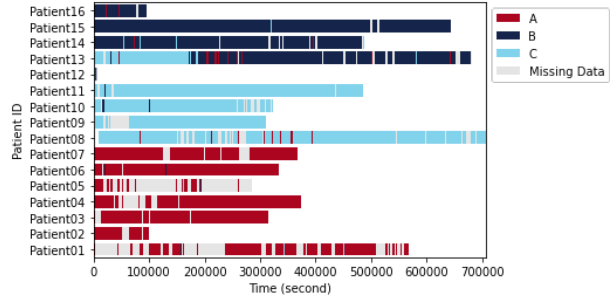

**Figure 2.** Physiological states of TBI patients over time

*Comparison of Clinical Variables across Physiological States*
The three physiological states A, B, and C are characterized as follows:
- State A is characterized by a low RR, low SBP, low CPP, and high ICP. This combination of parameters may indicate inadequate perfusion in the setting of intracranial hypertension.
- State B is characterized by a high RR, low $SpO_2$, high HR, and low ICP. These parameters suggest a state of respiratory distress or failure, such as pneumonia.
- State C is characterized by normal $SpO_2$, high SBP, high CPP, and low HR. These parameters suggest a state of relative cardiovascular and respiratory stability. This state implies that the patient is in a relatively stable condition.

There is a statistically significant difference between physiological states A, B, and C in all multivariate time-series clinical variables ($p < 0.001$) as demonstrated Figure 2. While the Kruskal-Wallis test indicates a statistically significant difference among the three physiological states concerning PRx and DBP, the lack of clear distinction in their plots suggests that these variables may not be as distinctive as the other multivariate time-series variables.

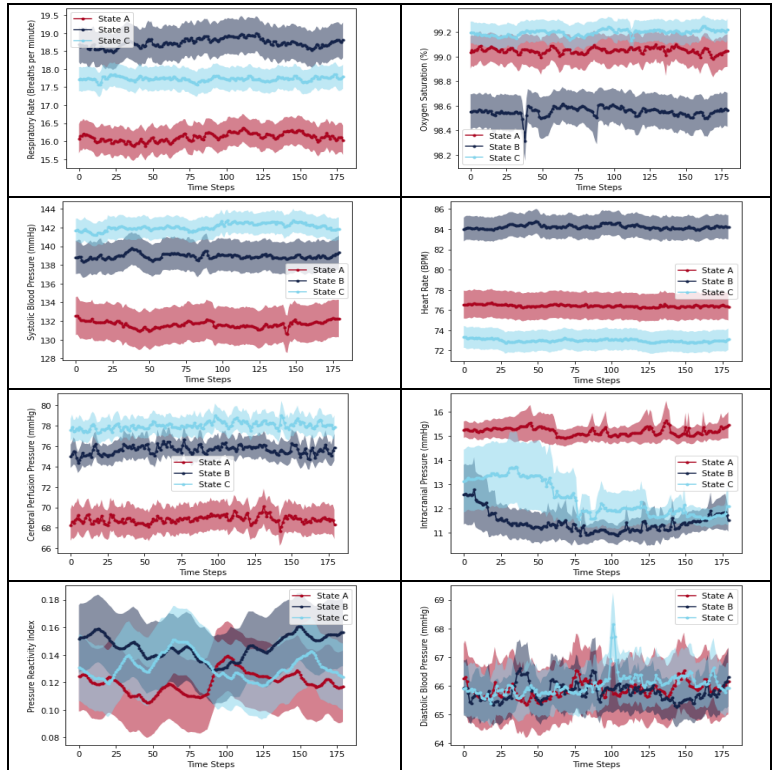

**Figure 3.** Comparison of distinctive time-series clinical variables across three physiological states

*Comparison of Patient Outcomes across Physiological States*

State A is the predominant state observed among the patients, with seven out of 16 patients spending more than 95% of their monitoring time in this state. This state is associated with a high risk of adverse outcomes, as evidenced by the fact that five of these patients died, one had a severe disability, and one had a moderate disability. On the other hand, state B was the least commonly observed state among the patients, and it was the predominant state in only three out of 16 patients. State C was observed as the predominant state in six out of 16 patients, with a Glasgow Outcome Scale Extended (GOSE) score range of 1 to 8. Among those patients, one had a GOSE score of 8, indicating a favorable outcome. However, four of those patients ultimately died. This suggests that while state C may have a lower risk of adverse outcomes compared to state A, it may still involve the risk of critical care related complications, particularly if the underlying neurological condition is severe.

**Table 2.** Frequencies of physiological states in TBI patients

| Patient IDs* | Frequency of State A | Frequency of State B | Frequency of State C |
|---|---|---|---|
| Patient 1 | 175 (98.3%) | 1 (0.6%) | 2 (1.1%) |
| Patient 2 | 45 (100%) | 0 (0%) | 0 (0%) |
| Patient 3 | 163 (100%) | 0 (0%) | 0 (0%) |
| Patient 4 | 174 (100%) | 0 (0%) | 0 (0%) |
| Patient 5 | 38 (95%) | 1 (2.5%) | 1 (2.5%) |
| Patient 6 | 180 (99%) | 2 (1%) | 0 (0%) |
| Patient 7 | 184 (100%) | 0 (0%) | 0 (0%) |
| Patient 8 | 10 (2.8%) | 0 (0%) | 342 (97.2%) |
| Patient 9 | 0 (0%) | 0 (0%) | 148 (100%) |
| Patient 10 | 0 (0%) | 3 (1.8%) | 161 (98.2%) |
| Patient 11 | 0 (0%) | 1 (0.4%) | 264 (99.6%) |
| Patient 12 | 1 (25%) | 0 (0%) | 3(75%) |
| Patient 13 | 12 (3.4%) | 250 (69.8%) | 96 (26.8%) |
| Patient 14 | 2 (0.8%) | 253 (97.7%) | 4 (1.5%) |
| Patient 15 | 1 (0.3%) | 351 (99.4%) | 1 (0.3%) |
| Patient 16 | 0 (0%) | 2 (4%) | 48 (96%) |

*Analysis of Impact of Clinical Events on Physiological States in TBI Patients*
Here, we explore the effects of clinical events on the physiological states of TBI patients. Our analysis focuses on six out of the 16 TBI patient records that displayed changes in their physiological states and had documented clinical events. These patients include Patient 1, Patient 5, Patient 8, Patient 13, Patient 14, and Patient 15. To investigate the impact of clinical events on the physiological states of these TBI patients, we divide them into four distinct groups based on their predominant physiological states. Patients 1 and 5 primarily exhibit Physiological State A, while Patients 14 and 15 have State B as their dominant physiological state. Patient 8 mainly presents with Physiological State C, and Patient 13 displays a combination of Physiological States B and C as the prevailing states.

*Impact of Clinical Events on Physiological State A:* In Patient 1, travel (i.e., movement within the ICU) led to a short-term transition from State A to State B, potentially due to the onset of respiratory distress or failure (Figure 4a). The physical movements associated with travel might have temporarily increased the patient's ICP, worsening their condition. On the other hand, suctioning procedures resulted in a shift from State A to State C. This change is likely due to improved airway clearance and oxygenation, which in turn lowered ICP and increased CPP. Suctioning might have also alleviated any respiratory distress, causing a decrease in the RR and an increase in HR.

For Patient 5, administering norepinephrine and hypertonic saline appears to have caused a change in the physiological state from State A to State C (Figure 4b). This is evidenced by increased blood pressure, CPP, and SpO$_2$, as well as a decrease in ICP and HR. Furthermore, introducing dihydroergotamine and acetazolamide likely prompted a shift from State A to State B. This change is characterized by reduced CPP, ICP, and SBP, along with elevated RR and HR, which are consistent with respiratory distress or failure.

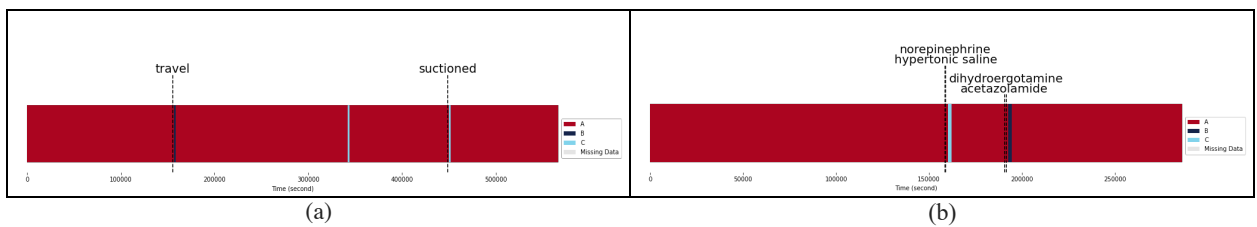

(a)      (b)

**Figure 4.** Physiological states of Patient 1 (a) and Patient 5 (b) over time with associated clinical events

*Impact of Clinical Events on Physiological State B:* In case of Patient 14, increasing the fraction of inspired oxygen during an $FiO_2$ challenge led to a temporary shift from State B to State C (Figure 5a). This change can be attributed to improved respiratory and cardiovascular status resulting from the increased oxygen supply. Similarly, administering pressors or inotropes, which raise blood pressure and/or HR, prompted a transition from State B to State A. This change could be due to the increased ICP caused by elevated blood pressure and HR. Other events, such as patient movement, bathing, and neurological examinations involving head and neck manipulation, also caused a shift from State B to State C. These changes might be associated with temporary alterations in cerebral blood flow or ICP. Moreover, airway clearance during suction or airway procedures may cause transient changes in respiratory status. Lastly, increased sedation or analgesia indicates additional pain management, potentially leading to improved respiratory and cardiovascular status.

For Patient 15, a transcranial Doppler ultrasound (TCD) test prompted a temporary change from State B to State A (Figure 5b). This transition could be due to stimulation-related increases in ICP resulting from the TCD test.

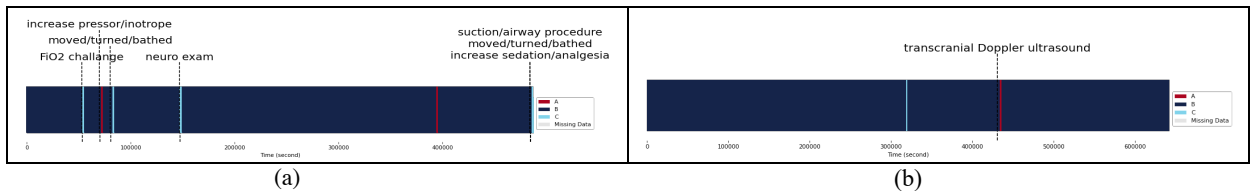

(a)          (b)

**Figure 5.** Physiological states of patient 14 (a) and patient 15 (b) over time with associated clinical events

*Impact of Clinical Events on Physiological State C:* In Patient 8, State C is the prevailing physiological state; however, various clinical events trigger temporary transitions to State B at different times (Figure 6). One such event includes the malfunction of an arterial line, which can produce inaccurate blood pressure readings. This inaccuracy may result in improper blood pressure management, causing a shift from State C to State A or may result in faulty state definitions, suggesting a need for artifact reduction techniques. Temporary discontinuation of hypertonic solutions, combined with the presence of diabetes insipidus, may lead to dehydration and excessive urination. These factors contribute to the transition from State C to State A. Moreover, sudden changes in sedation or analgesia levels can provoke a rebound effect, characterized by a rapid drop in blood pressure and an increase in ICP. This effect triggers a shift from State C to State A. Utilizing the Arctic Sun system to rapidly lower body temperature in order to manage elevated ICP can also cause a reduction in blood pressure and CPP. This change contributes to the transition from State C to State A. Additionally, ceasing the use of compression boots abruptly may lead to increased blood flow in the legs, causing stimulation and resulting in a change from State C to State A. Administering medication via a nasogastric tube may cause aspiration, leading to respiratory distress and decreased $SpO_2$ levels. This situation contributes to the shift from State C to State A. Furthermore, administering a bolus of sedation or analgesia can cause a sudden drop in blood pressure and an increase in ICP, leading to a change from State C to State A. Finally, interventions such as increasing the RR to 14 and administering a bolus of sedation might also contribute to the transition from State C to State A.

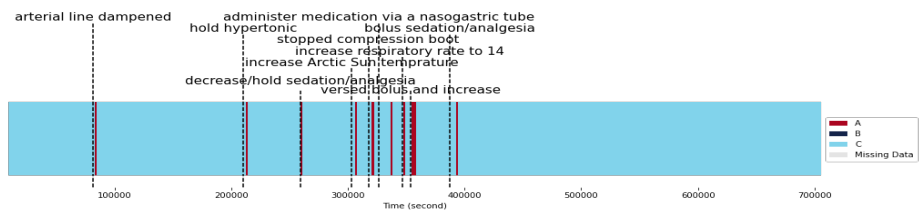

**Figure 6.** Physiological states of patient 8 over time with associated clinical events

*Impact of Clinical Events on Physiological States B and C:* Patient 13 primarily exhibited two physiological states, B and C, which were influenced by a variety of clinical events (Figure 7). Initially, the patient experienced State C, which was impacted by interventions such as the use of a clavicle guard/brace stimulation associated with placement of a mastoid needle electrode. The physiological changes resulting from these interventions could be related to the patient's discomfort or pain, potentially triggering a stress response and leading to a transition from State C to State

B. Brief disruptions in monitoring or adjustments in patient positioning while disconnecting the patient from the CT scan machine may cause temporary shifts from State C to State A, temporarily affecting ICP or other physiological parameters.

Following a procedure, Patient 13 underwent a change from State C to State B that was not temporary. This change could have been induced by the procedure itself, causing respiratory distress or increased ICP, or it could have been influenced by unrelated factors, such as infection or seizure. Activities like moving, turning, or bathing the patient may result in temporary shifts from State C to State B, characterized by increased HR and RR. However, reducing pressor/inotrope medication may stabilize the patient's physiological state and lead to a subsequent return to State C.

Temporary shifts from State B to State A can result from administering bolus sedation/analgesia and performing suction/airway procedures, as they increase HR and RR. However, administering mannitol or hypertonic saline can help reverse this change by reducing cerebral edema and increasing CPP, leading back to State C. In a separate instance, mannitol or hypertonic saline administration facilitated a transition from State B to State C through the same mechanism. By increasing pressor/inotrope medication, the physiological state can be stabilized, and the patient can return to State C, counteracting the temporary increase in HR and RR induced by other interventions. Although suction/airway procedures might cause temporary changes from State B to State A due to increased HR and RR, this effect may resolve on its own, leading to a subsequent return to State C.

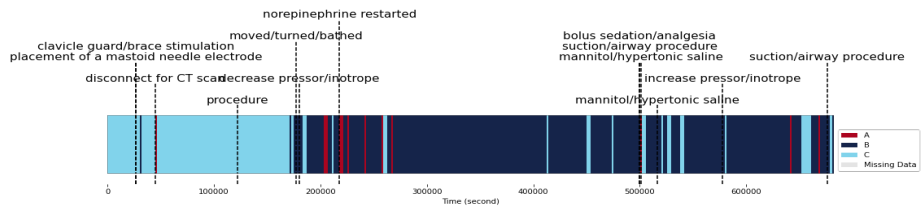

**Figure 7.** Physiological states of patient 13 over time with associated clinical events

**Discussion**

Using SLAC-Time to cluster multivariate time-series physiological data from TBI patients, we identified three unique physiological states: A, B, and C. State A is a critical state resulting from TBI, marked by inadequate oxygen and blood supply to the brain resulting from low CPP and high ICP. Immediate interventions, such as mechanical ventilation, vasopressors, or surgery, may be necessary to reduce ICP and stabilize vital signs. In addition, administering mannitol or hypertonic saline in State A patients can help reduce cerebral edema and increase CPP. Additionally, administering norepinephrine and hypertonic saline may also be effective in treating patients in State A, as these medications can increase blood pressure, CPP, and SpO$_2$, while reducing ICP and HR.

State B suggests a state of respiratory distress, where the patient is breathing rapidly but not getting enough oxygen, leading to low SpO$_2$ and high HR. State B may indicate an initial stage of injury where the body is compensating for reduced brain perfusion by increasing breathing and HR, or compromised oxygen supply to the brain related to pulmonary pathology such as pneumonia. In this regard, suctioning may be effective in aiding airway clearance, and oxygenation, and alleviating respiratory distress. In addition, increasing the fraction of inspired oxygen can improve respiratory and cardiovascular status in TBI patients in state B. Oxygen supplementation is recommended to maintain adequate SpO$_2$ levels in TBI patients. Increasing sedation and analgesia may also be beneficial in improving respiratory and cardiovascular status for TBI patients in state B. Adequate pain control and sedation are recommended in part to improve respiratory and cardiovascular status in TBI patients.

State C represents a relatively stable state in which brain is receiving adequate oxygen and perfusion in the setting of optimized hemodynamics. However, the relatively low HR may indicate potential bradycardia, which needs further investigation.

*Limitations*

This study has several limitations that should be taken into consideration. First, our analysis was limited to only 16 TBI patient records, which may not be sufficient to fully identify TBI physiological states. Future studies may benefit from including a larger sample size. Second, the study was conducted using data from a single center within the TRACK-TBI dataset, which limits the generalizability of the identified TBI physiological states. Further validation against other datasets is necessary to establish their applicability. Third, our study was limited to only eight clinical variables. Incorporating additional clinical variables may provide further insights into physiological states. Moreover, our analysis of the impact of clinical events on physiological states was limited to six patients. A larger sample size may be necessary to validate our findings on the effects of clinical events on each physiological state.

**Conclusion**
Our study identified three distinct physiological states of TBI patients based on clinical time-series data using the SLAC-Time approach without imputing or aggregating missing data. The identified physiological states (A, B, and C) showed significant differences in all multivariate time-series clinical variables, with State A being the predominant state associated with a high risk of adverse outcomes. These findings have important clinical implications for the development of personalized treatment strategies based on the specific physiological state of TBI patients, which could lead to improved clinical outcomes. To further advance research in this area, future studies should consider increasing the sample size, incorporating additional clinical variables and events, and comparing results from the TRACK-TBI dataset to other datasets to improve generalizability.

**Acknowledgments**
The research presented in this paper has been made possible through funding provided by the National Science Foundation via grants #1838730 and #1838745. Dr. Foreman received financial support from the National Institute of Neurological Disorders and Stroke, part of the National Institutes of Health, under the grant K23NS101123. The interpretations, views, and recommendations provided in this work are the sole responsibility of the authors and do not necessarily align with those of the National Science Foundation or the National Institutes of Health. We would like to extend our gratitude to the investigators of the TRACK-TBI Study for granting us access to the data utilized in this research.